# DC and SA: Robust and Efficient Hyperparameter Optimization of Multi-subnetwork Deep Learning Models

Alex H. Treacher, Albert Montillo,
for the Alzheimer's Disease Neuroimaging Initiative*

Lyda Hill Department of Bioinformatics
University of Texas Southwestern Medical Center
Dallas, Texas, USA

**Abstract:**

We present two novel hyperparameter optimization strategies for optimization of deep learning models with a modular architecture constructed of multiple subnetworks. As complex networks with multiple subnetworks become more frequently applied in machine learning, hyperparameter optimization methods are required to efficiently optimize their hyperparameters. Existing hyperparameter searches are general, and can be used to optimize such networks, however, by exploiting the multi-subnetwork architecture, these searches can be sped up substantially. The proposed methods offer faster convergence to a better-performing final model. To demonstrate this, we propose 2 independent approaches to enhance these prior algorithms: 1) a divide-and-conquer approach, in which the best subnetworks of top-performing models are combined, allowing for more rapid sampling of the hyperparameter search space. 2) A subnetwork adaptive approach that distributes computational resources based on the importance of each subnetwork, allowing more intelligent resource allocation. These approaches can be flexibily applied to many hyperparameter optimization algorithms. To illustrate this, we combine our approaches with the commonly-used Bayesian optimization method. Our approaches are then tested against both synthetic examples and real-world examples and applied to multiple network types including convolutional neural networks and dense feed forward neural networks. Our approaches show an increased optimization efficiency of up to 23.62x, and a final performance boost of up to 3.5% accuracy for classification and 4.4 MSE for regression, when compared to comparable BO approach.

Index terms – hyperparameter optimization, deep learning, subnetwork, multi-input, multi-modality.

* Data used in preparation of this article were obtained from the Alzheimer's Disease Neuroimaging Initiative (ADNI) database (adni.loni.usc.edu). As such, the investigators within the ADNI contributed to the design and implementation of ADNI and/or provided data but did not participate in analysis or writing of this report. A complete listing of ADNI investigators can be found at: http://adni.loni.usc.edu/wp-content/uploads/how_to_apply/ADNI_Acknowledgement_List.pdf

# 1. Introduction and background

Performance of deep learning models is problem specific and highly dependent on the selection of hyperparameters that define the model's configuration (e.g. number of hidden layers in the network, learning rate, activation function, etc.). To maximize model performance for each task, unbiased and efficient optimization strategies are required. The main goal of hyperparameter optimization (HPO) is to find an optimal model configuration to maximize predictive performance, while simultaneously minimizing the computational resources required to do so. Many methods for HPO have been suggested. Early methods focused on more stochastic optimization including hand tuning [1] and random search [2]. Later work often focused on one of 3 things to increase the efficacy of HPO: 1) network modification, 2) intelligent selection of hyperparameters, and 3) partial training. Initial work in network modification included Net2Net [3] and population based training [4]. These methods broadly focus on modifying trained networks to efficiently test novel model configurations without training each configuration de novo. Intelligent hyperparameter (HP) sampling started with the application of Bayesian optimization (BO) [5], which uses past knowledge of the hyperparameter space based on trained configurations to sample subsequent configurations that are more likely to provide higher relative performance. Lastly, successive halving and hyperband [6] introduced the idea of partial training of networks, allowing computational resources to be more efficiently applied to models more likely to perform well, while underperforming models are stopped early. Many following works improve on these ideas and/or combine elements from all three. For example, NAS [7], ENAS [8] and PNAS [8] use intelligent HP selection along with network modification, and BOHB [9] combines intelligent HP selection with the partial training of hyperband.

While many hyperparameter optimization methods have been developed, most existing methods are general approaches and therefore agnostic to overall network architecture. More complex deep learning models that exhibit multi-subnetwork architecture are recently rising in use (fig. 1). Multi-subnetwork models consist of multiple smaller networks, which are combined into a single model for a specific task or set of tasks. Deep learning networks for multi-modal input data often employ subnetwork architecture such that each input has a corresponding dedicated subnetwork. Such networks have been successfully employed for word recognition from a combination of audio and video inputs [10], prediction of neurodegenerative diagnosis from multi-modal imaging [11], personality prediction from audio and video [12], emotion recognition from EEG and eye motion data [13], drug response from multi-omics data combined with multiple encoding networks [14], and multi-modal fusion of earth observation data (e.g. light detection and ranging, synthetic aperture radar, hyperspectral imagery, etc.) for land classification (e.g. grass, trees soil, water, etc) [15]. Other work has shown success using subnetwork architectures that combine multiple subnetworks of different types, often to a single modality of data. Zhao et al. [16] used a subnetwork model consisting of a fully convolutional neural network and a conditional random field recurrent neural network to predict brain tumor segmentation from multi-modal MRI imaging. Wu et al. [17] combined a UNet-like architecture with 2 CNN subnetworks to both segment the carotid artery vessel wall and diagnose carotid atherosclerosis from T1 weighted MRI. Gu et al. [18], used a two subnetwork architecture for smoke detection from a still image. Lastly, Chiu et al. [19] predicted predict drug response from mutation and expression profiles a cancer or tumor cells, using a model that combined encoder/decoder subnetworks for each input, along with a dense feed forward predictor network.

As the demand for multi-subnetwork neural networks increases, so does the need for efficient HPO methods. Very few related works on HPO of models with subnetworks have been published. MFAS [20] defines an efficient method to merge two independent, previously trained, single input, networks into a multi-subnetwork model, but the subnetworks are fixed. Zhou

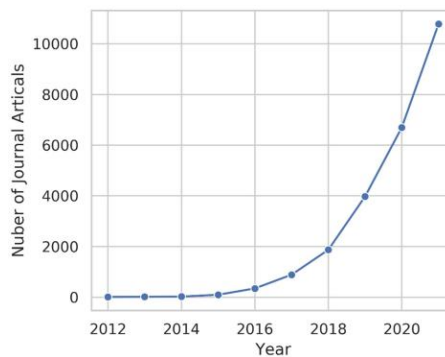

**Fig 1**. Number of manuscripts per year relating to multi-subnetwork deep learning. Manuscripts found using the Scopus search: TITLE-ABS-KEY ( "deep learning" AND ( "multimodal" OR "subnetwork" ) ) AND ( LIMIT-TO ( DOCTYPE,"ar" ) )

Yu et al. [21] describe a method for multimodal HPO that does allow for modification of the subnetworks. Their method focuses on building neural networks that have two inputs, image and text. They optimize a "backbone" model that offers high performance across multiple text/image tasks, and they produce task-specific head models that take the output from the backbone and predict for the specific task. Our work offers new, more general, optimization strategies for the optimization of a multi-modal network that overcome restrictions of the other multi-modal optimization strategies. Unlike the previous methods, the number of inputs is not limited to two, the subnetworks are also optimized in conjunction with the merge network, training on multiple related tasks is not required, and there is no restriction on the input type.

In this work we provide 2 novel approaches, divide-and-conquer (DC) and subnetwork adaptive (SA) HPO, capable of enhancing existing hyperparameter optimization methods when applied to networks that exhibit multi-subnetwork architecture. We apply these approaches to the commonly-used BO method for HPO, creating divide-and-conquer Bayesian optimization (DCBO) and subnetwork adaptive Bayesian optimization (SABO), and demonstrate increased efficiency. To our knowledge, these are the first modality-independent and complete HPO algorithms for multi-input neural networks. These methods are tested on both a CIFAR-based semi-synthetic dataset, which allows us to test the algorithms with different experimental parameters, including number of inputs, and amount of training data, as well as the prediction of 3-year cognitive trajectory in patients with mild cognitive impairment (MCI) and Alzheimer's disease (AD). The proposed algorithms consistently increase the efficiency of the HPO and providing a speedup of up to 23.62x and a final performance boost of up to 3.5% accuracy for classification and 4.4 MSE for regression when compared to BO.

## 2. Methods and Algorithms

### 2.1 Summary of Bayesian optimization and Tree Parzen Estimator (TPE).

Bayesian optimization (BO) is a commonly-used and effective optimization strategy for black-box functions, making it an appropriate method for hyperparameter optimization [5, 22]. Suppose that the objective function $f(c) = l$ represents a model trained with configuration $c$ resulting in a loss $l$. BO uses a probabilistic model $p(f|C^T, L^T)$ to estimate a probability density of the objective function $f$ based on $j$ prior point observations of models configurations $C^T = \{c_1, c_2, \dots c_j\}$ and corresponding losses $L^T = \{l_1, l_2 \dots l_j\}$. An acquisition function is then fit based on the probabilistic model $p$, and maximized to select a new configuration ($c^*$) to test. $f(c^*)$ is then evaluated and $l^*$ calculated, then appended to $C^T$ and $L^T$, respectively. This iterative process then restarts by refitting the probabilistic model $p$ and acquisition function, based on the updated $C^T$ and $L^T$. In this fashion, $f$ is optimized without requiring a closed form equation. This work uses Tree Parzen Estimators (TPEs) [5] to define $p$ (eq. 3) and the acquisition function (eq 4). TPE is an efficient form of BO for optimization over high dimensional spaces where configuration selection scales linearly with the number of models trained, as opposed to Gaussian Processes that scale cubically [5]. Kernel density estimators are used to model the probability density of the high (eq 1) and low (eq 2) performing networks, based on the top performing $\alpha$ percentile, such that, $c^*$ is then selected (eq. 4) for the next selected configuration.

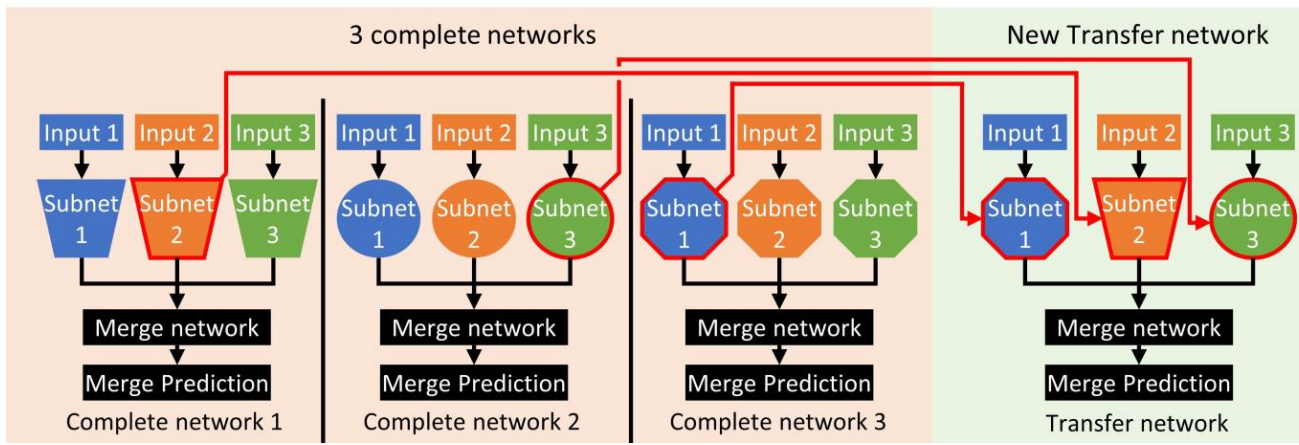

**Fig 2**: Creation of a transfer network. $I$ (where $I$ is the number of inputs in a network) networks can be combined to create a new transfer model. Both the configuration and weights can be transferred. This allows for the creation of a unique and mostly trained transfer model.

$$\ell(c) = p(l < \alpha | c, C^T, L^T) \quad (1)$$

$$g(c) = p(l > \alpha | c, C^T, L^T) \quad (2)$$

$$p(c|l) = \begin{cases} \ell(c) \; if \; l < l^* \\ g(c) \; if \; l \geq l^* \end{cases} \quad (3)$$

$$c^* = argmax\left(\frac{\ell(c)}{g(c)}\right) \quad (4)$$

**2.2 Divide and conquer Bayesian optimization (DCBO).**

The first of two novel methods in this work, the divide-and-conquer (DC) approach, leverages the separability of the subnetworks to more efficiently train networks that contain multiple subnetworks. To achieve this, entire neural networks are trained from random initialization (henceforth referred to as *complete* networks), then individual subnetworks from high-performing? complete networks are combined using transfer learning to create a new network with a novel configuration ($c^*$) (henceforth referred to as *transfer* networks) (fig. 2). Training transfer networks provides 2 major benefits over training complete networks: First, as performance of the network is dependent on all subnetworks, combining relatively high-performing trained subnetworks, from different complete models, will likely yield a transfer model that out performs any of the complete models it was created from. And second, as the majority of the weights in the transfer model are transferred from a trained complete model, the transferred subnetwork weights can be frozen and therefore greatly reduce the computational resources required to train the transfer network. These advantages allows for a relatively much larger number of transfer models to be tested compared to training complete models (as other optimization strategies use), with the same amount of computational resources. The combinatorics of combining different parts of complete models ensures a large diversity in the possible transfer networks. For example, to build each transfer model, up to $I$ complete models are combined, where $I$ is the number of subnetworks. Training 1000 complete networks, from random initialization would take $1000\bar{r}_c$, where $\bar{r}_c$ is the average amount of resources to train a complete network. With $I = 3$, only 10 complete networks are needed to create $10^3 - 10 \cong 10^3$ different and unique transfer networks. In general, the divide and conquer approach can provide a maximum speed up of $T^{I-1}$ (eq. 5).

$$\frac{T^I R}{TR} = T^{I-1} \quad (5)$$

It's worth noting that with a single subnetwork ($I = 1$) the max speed up is $T^0 = 1$. This makes sense as with a single subnetwork, DC becomes equivalent to the general optimization strategy. Compared to existing methods that train complete models, DC provides the potential for a speed up that grows exponentially with the number of subnetworks ($I$). In reality, the transfer networks will likely require some fine-tuning and hence the actual speed up will be attenuated by the fraction of time required to fine-tune the transfer network. For multi-input models, this is involves training a relatively small merge network that combines the outputs of each subnetwork and produces the final prediction. Lastly, we do not recommend testing all combinations. As described below, subnetworks that are more likely to provide a high relative performance are combined, rather than all combinations. This avoids training models that are likely to have poorer performance.

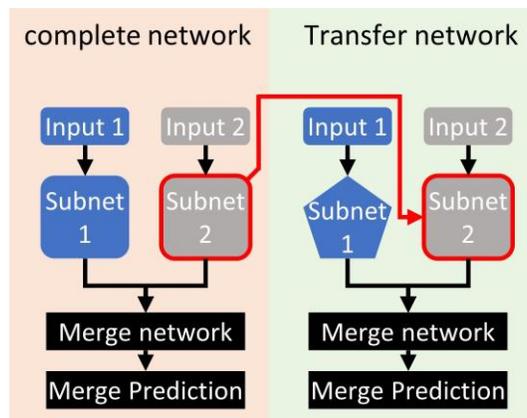

**Fig 3**: Creation of a transfer network in SABO. Less important input and subnetwork shown in grey. Both the configuration and weights can be transferred with a probability inversely proportional to the input's importance. This allows for the creation of new models that spend more computational resources on the more important subnetwork. In our examples, importance is defined by relative performance across the subnetworks for each input.

To demonstrate the speedup over the corresponding general HPO methods, DC is combined with BO to create DCBO. The DCBO algorithm (algorithm 1) describes one iteration of model selection and training given the previously trained models configurations ($C^T$), losses ($L^T$), and model weights ($\Theta^T$). This algorithm is the iterated, each time selecting and testing a new model configuration ($c^*$) to acquire the corresponding model loss ($l^*$), and model weights ($\theta^*$). These are then appended to $C^T, L^T$ and $\Theta^T$, respectively, to provide updated $C^{T*}, L^{T*}$ and $\Theta^{T*}$. This is iterated until the computational resource limit ($R$) is reached. Prior to training any models, $C^T, L^T$, and $\Theta^T$ are empty sets. Briefly, lines 1-25 are used to select hyperparameters, and transfer weights from complete models when appropriate. The exact method used to select $c^*$ depends on the availability of complete networks, as well as the ability to create TPEs. TPEs require $N+1$ samples, where $N$ is the number of hyperparameters to optimize. As with commonly used BO, to help with exploration, there is a $v$ probability a model's configuration is chosen at random, and here it is set to 1/3. Also, to ensure that there are ample models to transfer from, $o$ defines the probability that weights are trained from random initialization (aka complete model), which we set to N. This means, that the of number complete models are trained prior to transfer learning scales with the number of hyperparameters. This helps to ensure ample sampling of the hyperparameters space prior to the application of BO.

## 2.3 Subnetwork Adaptive BO.

The second method this work presents is subnetwork adaptive (SA) HPO. Existing approaches do not take into consideration that subnetworks may differ in importance for overall model performance, and the importance of each subnetwork may be unknown at the time of training. We propose spending more computational resources optimizing the parts of the network that have a larger impact on the overall performance. To do so for multi-input neural networks, we adaptively weigh the probability that an input's subnetwork will be transferred from the top performing subnetworks for that input. When subnetworks from a specific input are performing poorly, rather than train the new configurations from random initial weights, we transfer the best trained subnetwork so far, therefore greatly reducing the computational budget used for this input. (fig. 3) Similarly to DCBO, we combine out SA method with BO to create SABO.

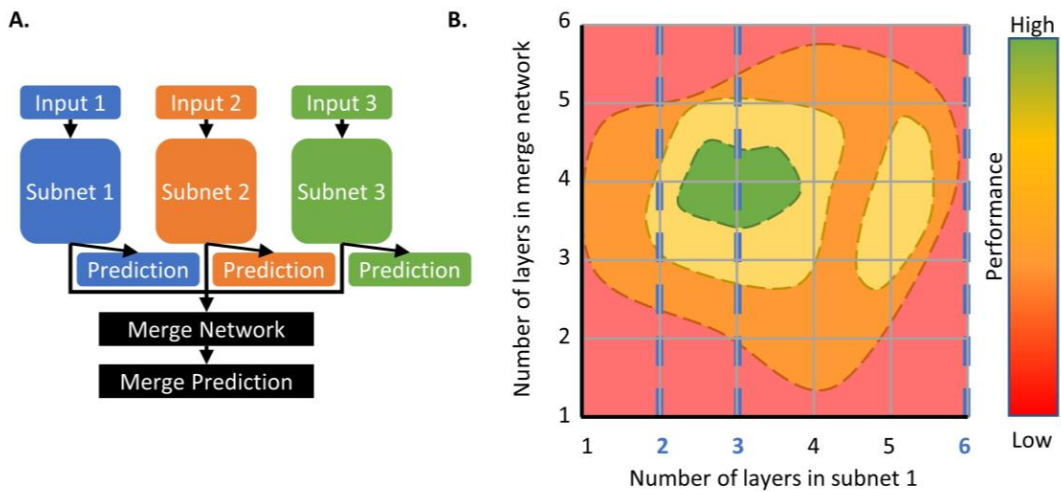

**Fig 4.** Technical developments to allow for the DC and AI methods, and their combination with BO **A.** Modifications to the model, with an updated loss function (eq 4). **B.** Focal TPE. A simplified visualization of focal TPE on a 2D hyperparameter space. On the x axis is the number of layers in subnet 1, and on the y axis the number of layers in the merge network. The unknown performance of the hyperparameter space is indicated by the background color. Hyperparameters of previously trained models are indicated by the blue numbers and corresponding vertical lines for subnet 1. While the KDE's for the TPE covers the entire space, the acquisition function is strictly limited to search along the blue lines for hyperparameters in subnet 1, but has freedom to select any value for the hyperparameters for the merge network (y axis). This allows for any of the hyperparameters to be set to specific discrete values, required for using TPE in conjunction with transfer learning.

**Table 1**: Summary of experiments included in this work. The letters in the subscript of the abbreviation are method (D=DCBO,A=SABO), dataset (C=CIFAR,A=ADNI), Class (C=Classification, R=Regression), Model (C=CNN,D=DFNN), respectively. For the ADNI datasets, data size is "the number of unique subjects(number of unique scans)".

| Experiment Abbreviation | Method | Dataset | Class | Target | Model | Datasize |
|---|---|---|---|---|---|---|
| $E_{DCCC}$ | DCBO | CIFAR | Classification | Image Class | CNN | 60,000 |
| $E_{DARD}$ | DCBO | ADNI | Regression | Future ADAS-cog | DFNN | 287(353) |
| $E_{DACD}$ | DCBO | ADNI | Classification | MCIs vs MCIc | DFNN | 330(425) |
| $E_{DACC}$ | DCBO | ADNI | Regression | Future ADAS-cog | CNN | 315(399) |
| $E_{ACCC}$ | SABO | CIFAR | Classification | Image Class | CNN | 60,000 |
| $E_{AARD}$ | SABO | ADNI | Regression | Future ADAS-cog | DFNN | 398(571) |
| $E_{AACD}$ | SABO | ADNI | Classification | MCIs vs MCIc | DFNN | 472(837) |

SABO (Algorithm 2), is also an iterative process. However the relative performance of each input is calculated using the mean 90$^{th}$ percentile performance per fold over all subnetworks trained for that input (Algorithm 2, line 2), and used to determine the probability of applying transfer learning from a complete model. Consequently, transfer learning is more likely to be applied to inputs with lower performance. The configuration is selected (Algorithm 2, lines 3-22) for the new model. The selection uses TPEs (Algorithm 2, line 19), and transfer learning (Algorithm 2, line 10, 22) when available.

**2.4 Technical developments to support the DC and SA methods:**

Both DCBO and SABO require relative performance measures $l$ for each of the subnetworks. In order to provide an estimate of subnetwork performance, a small prediction layer added to each subnetwork, and an additional loss term for each subnetwork is added (Fig. 4a, eq. 6).

$$l = l_M + \Lambda \sum_i^I l_i \qquad (6)$$

In addition to the model modification, we developed focal TPE (Fig. 4b, eq 7), which is an adaptation of TPE that allows transfer learning to work in conjunction with BO. To achieve this, focal TPE replaces the acquisition function (eq. 3) with equation 7 and restricts the search space for subnetwork hyperparameters to points that have already been sampled. The use of focal TPE not only allows for flexible selection of hyperparameters unrelated to transfer learning ($c_m$), such as the number of layers in the merge network, but also inherently allows BO to optimize the combination of trained subnetworks ($C_1^T, \ldots C_I^T$) for each subnetwork from different complete networks.

$$c^* = argmax \left( \frac{l(c=\{c_1 \in C_1^T, c_2 \in C_2^T \ldots c_I \in C_I^T, c_m\})}{g(c=\{c_1 \in C_1^T, c_2 \in C_2^T \ldots c_I \in C_I^T, c_m\})} \right) \qquad (7)$$

**2.5 Implementation**

DCBO and SABO are written in python v3.7.7 [23], and modified from HpBandStr. Keras 2.2.4 [24] and Tensorflow 1.13.1 [25] are used to define and train machine learning models. BO was implemented with HpBandStr and ConfigSpace 0.4.13. The code base makes further use of numpy 1.18.1 [26], pandas 1.0.3 [27], sklearn 1.1.dev0 and 0.22.1 [28], and GNU parallel 20150122 [29]. Imaging processing pipelines (as described in materials) used Freesurfer 7.1.1 [30], FSL v5.0.10 [31], and ray v1.2.0 [32].

| Algorithm 1: DCBO | Algorithm 2: SABO |
|---|---|
| *Inputs:* $C^T, L^T, L_I^T, C, \Theta, v, o, X, Y, L$ | *Inputs:* $C^T, L^T, L_I^T, C, \Theta, v, o, r, R, X, Y, L$ |
| *Output:* $C^{T*}, L^{T*}, L_I^{T*}, M^{T*}$ | *Output:* $C^{T*}, L^{T*}, L_I^{T*}, M^{T*}$ |
| 1. If $\|C_i^T\| \leq dim(C_i)$ or $(r() < v$ and $r() < o)$ | 1. for $i$ in $I$: |
| 2. $\quad c^* \in_R C$ | 2. $\quad p_i = \frac{mean(L_i^{90th})}{\Sigma_i^I mean(L_i^{90th})}$ |
| 3. If $\|C_i^T\| \leq dim(C_i)$ and $r() < v$ and $r() > o$ | 3. if $\|C_i^T\| \leq dim(C)$ or $(r() < v$ and $r() < o)$: |
| 4. $\quad$ For $i$ in $I$: | 4. $\quad c^* \in_R C$ |
| 5. $\quad\quad c_i^* \in_R C_i^T$ | 5. if $\|C^T\| \leq dim(C)$ and $r > v$ and $r() > o$: |
| 6. $\quad\quad \theta_i = M_i^T(c_i^*)$ | 6. $\quad$ For $i$ in $I$: |
| 7. $\quad\quad Freeze(\theta_i)$ | 7. $\quad\quad$ if $p_i < r()$: |
| 8. $\quad c_m^* \in_R C_m$ | 8. $\quad\quad\quad c_i^* \in_R C_i^T$ |
| 9. if $\|C_i^T\| > dim(C_i)$ and $\|C^T\| < dim(C)$ and $r() > v$ and $r() > o$ | 9. $\quad\quad\quad \theta_i = M_i^T(c_i)$ |
| 10. $\quad$ For $i$ in $I$: | 10. $\quad\quad\quad Freeze(\theta_I)$ |
| 11. $\quad\quad c_i^* = ResTPE_i(C_i\|C_i^T, L_i^T)$ | 11. $\quad\quad$ else: |
| 12. $\quad\quad \theta_i = M_i^T(c_i^*)$ | 12. $\quad\quad\quad c_i^* \in_R C_i$ |
| 13. $\quad\quad Freeze(\theta_i)$ | 13. $\quad c_m^* \in_R C_m$ |
| 14. $\quad c_m^* \in_R C_m$ | 14. if $\|C^T\| > dim(C)$ and $r > v$ and $r() > o$: |
| 15. if $\|C_i^T\| > dim(C_i)$ and $\|C^T\| < dim(C)$ and $r() > v$ and $r() < o$ | 15. $\quad C' = C$ |
| 16. $\quad$ For $i$ in $I$ | 16. $\quad$ For $i$ in $I$: |
| 17. $\quad\quad c_i^* = TPE_i(C_i\|C_i^T, L_i^T)$ | 17. $\quad\quad$ if $p_i < r()$: |
| 18. $\quad c_m^* \in_R C_m$ | 18. $\quad\quad\quad C_i' = C_i^T$ |
| 19. if $\|C^T\| > dim(C)$ and $r() > v$ and $r() > o$ | 19. $\quad c^*, \theta = ResTPE(C'\|C^T, L^T), L^T$ |
| 20. $\quad c^*, \theta = ResTPE(C\|C^T, L^T), M^T$ | 20. $\quad$ For $i$ in $I$: |
| 21. $\quad$ For $i$ in $I$: | 21. $\quad\quad \theta_i = M_i^T(c_i^*)$ |
| 22. $\quad\quad \theta_i = M_i^T(c_i^*)$ | 22. $\quad\quad Freeze(\theta_i)$ |
| 23. $\quad\quad Freeze(\theta_i)$ | 23. $init(\theta)$ |
| 24. if $\|C^T\| > dim(C)$ and $r() > v$ and $r() > o$ | 24. $\theta^* = argmin_\theta L(f(X, c^*, \theta), Y)$ |
| 25. $\quad c^* = TPE(C\|C^T, L^T)$ | 25. $C^{T*} = C^T \cup \{c\}$ |
| 26. $init(\theta)$ | 26. $L^{T*} = L^T \cup L(f_c(X, c, \theta^*), Y)$ |
| 27. $\theta^* = argmin_\theta L(f(X, c^*, \theta), Y)$ | 27. For $i$ in $I$: |
| 28. $C^{T*} = C^T \cup \{c^*\}$ | 28. $\quad L_i^{T*} = L_i^T \cup L_i(f_c(X, c^*, \theta^*), Y)$ |
| 29. $L^{T*} = L^T \cup L(f_c(X, c^*, \theta^*), Y)$ | 29. $\Theta^{T*} = \Theta^T \cup \{\theta^*\}$ |
| 30. For $i$ in $I$: | 30. Return $C^{T*}, L^{T*}, L_I^{T*}, \Theta^{T*}$ |
| 31. $\quad L_i^{T*} = L_i^T \cup L_i(f_c(X, c^*, \theta^*), Y)$ | |
| 32. $\Theta^{T*} = \Theta^T \cup \{\theta^*\}$ | |
| 33. Return $C^{T*}, L^{T*}, L_I^{T*}, \Theta^{T*}$ | |

**Table 2:** Speed up and performance gain summary of DCBO and ABO over BO. At each time point the speed up is calculated to be the minimum time BO took to reach it's current performance / minimum time DCBO took to reach the same performance as BO. *Mean values across the 5 different experiments (independent results in table 3)

| Experiment | Mean speed up | Max speed up | Final speed up | Final Gain in Performance |
|---|---|---|---|---|
| $E_{DCCC}$ | 4.81* | 9.84* | 8.30* | 3.1% (accuracy)* |
| $E_{DARD}$ | 15.27 | 23.62 | 22.84 | 3.62 (MSE) |
| $E_{DACD}$ | 2.52 | 5.72 | 1.81 | 1.1% (accuracy) |
| $E_{DACC}$ | 7.11 | 10.87 | 10.87 | 4.44 (MSE) |
| $E_{ACCC}$ | 2.28 | 4.08 | 3.24 | 1.8% (Accuracy) |
| $E_{AARD}$ | 4.44 | 6.48 | 6.48 | 3.22 (MSE) |
| $E_{AACD}$ | 2.85 | 3.66 | 3.66 | 1.1% (accuracy) |

# 3. Materials

## 3.1 CIFAR100.

CIFAR100 [33] is a dataset consisting of 50,000 training images and 10,000 validation images, each of size 32x32 pixels with 3 RGB color channels. Each image contains an object (e.g. car, dog, plane, apple) from 100 different classes. CIFAR is used for the construction of a pseudo-synthetic datasets as described for each of the experiments in their respective section.

## 3.2 Alzheimer's Disease Neuroimaging Initiative (ADNI) dataset.

ADNI [34] data used in this article were obtained from the Alzheimer's Disease Neuroimaging Initiative (ADNI) database (adni.loni.usc.edu). It was launched in 2003 lead by Michael W. Weiner MD. The primary goal of ADNI has been to test whether serial magnetic resonance imaging (MRI), positron emission tomography (PET), other biological markers, and clinical and neuropsychological assessment can be combined to measure the progression of mild cognitive impairment (MCI) and early Alzheimer's disease (AD). For up-to-date information, see www.adni-info.org.

DCBO and SABO are both applied to ADNI to provide a real word example of the utility of DCBO and SABO. There are 4 input modalities: clinical measurements, T1-weighted magnetic resonance imaging, positron emission tomography with fluorodeoxyglucose tracer (FDG-PET), and positron emission tomography with florbetapir tracer (AV45-PET). In these experiments, we focus on subjects with mild cognitive impairment (MCI) that have this data at a baseline date and a diagnosis and/or ADAS13 cognitive score 36 months into the future. Clinical data includes age, sex, years of education, and baseline scores for ADAS13, ADAS11, FAQ score, delayed recall total (LDELTOTAL), MMSE, MOCA, Rey's Auditory Verbal Learning Test (RAVLT) forgetting, RAVLT immediate, RAVLT learning, RAVLT percent forgetting, and trail making test (TRAB). We explore both a classification and a regression task. For classification, we predict if the MCI subject converts to AD (MCIc) or if they remain stable (MCIs). For regression, we predict the 36-month future ADAS13 score. A window of 90 days is allowed for the 36 month prediction, e.g. 36months+/-3months. Clinical, T1, and PET data was downloaded using

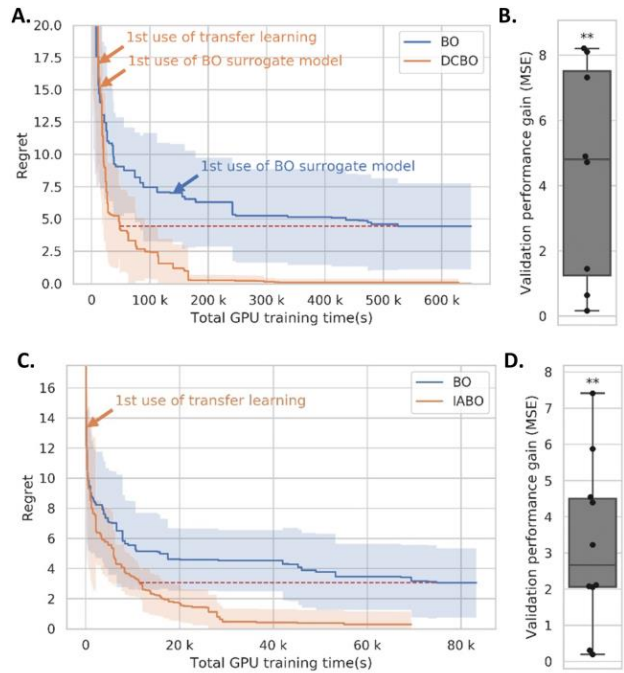

**Fig 5.** Regret plots as a function of GPU training time in seconds, and final performance gain (MSE) for comparing our methods to BO for experiments $E_{DARC}$ and $E_{AARD}$ respectively. Final speed up shown as a red dotted line **A.** Regret of DCBO vs BO. Average time for the first use of a surrogate function and transfer learning indicated for each method by arrows. **B.** Box plot showing the increase in validation performance per fold for the final model across the 8 folds. **C.** Regret of ABO vs BO. Average time for the first use of transfer learning indicated for each method by arrows. **D.** Box plot showing the increase in validation performance per fold for the final model across the 10 folds.

**Table 3**. Performance for DCBO experiments (E$_{DCCC}$) using CIFAR with different numbers of inputs, and different amount of training data.

| # of subnets | % data for training | Mean speed up | Max speed up | Final speed up | Final Gain in Performance |
|---|---|---|---|---|---|
| 3 | 20 | 5.42 | 15.78 | 7.74 | 3.2% |
| 4 | 20 | 5.63 | 8.75 | 8.51 | 2.7% |
| 9 | 20 | 2.67 | 4.00 | 3.99 | 3.5% |
| 4 | 50 | 5.30 | 8.54 | 8.54 | 3.2% |
| 4 | 100 | 3.80 | 7.49 | 7.49 | 3.3% |

IDA. Specifically, clinical data was taken from "ADNIMERGE". For PET imaging, the coregistered, averaged, and standardized images are used as inputs to the below pipelines.

To generate tabular data from the ADNI imaging data for the experiments using dense feedforward neural network (DFNN), the following steps are applied. 1) PET images are aligned to the subject's T1 image using FSL's rigid-body registration in FLIRT. 2) ConsNet [35] is used to skull-strip the T1, and 3) Freesurfer is used to spatially normalize and segment the T1 images. The Freesurfer segmentation is then used to segment the aligned PET image. Standardized uptake value ratios (SUVr) are computed for cortical and subcortical regions, normalized by the hemisphere-specific cerebellar mean uptake value. Additionally, mean volume is calculated for the subcortical (aseg), cortical (aparc), brainstem, amygdala, and hippocampal freesurfer segmentations of the T1 image.

An imaging processing pipeline was created for experiments using convolutional neural networks (CNNs): 1) PET to T1 rigid-body registration is done with ANTS. 2) ConsNet [35] is used to skull-strip the T1, and 3) Freesurfer is then used to spatially normalize and segment the T1 images. 4) The mean cerebellar uptake value is used to compute normalized SUVr maps for both FDG and AV45. Then, to ensure all subjects are orientated similarly, 4) the T1 for each subject is aligned to the AC-PC line of the MNI 152 standard brain [36], similarly using ANTS rigid-body registration. 5) The T1 images are downsampled to 3mm isotropic voxels, and the PET to 6mm isotropic voxels. This is required to fit GPU memory restrictions. For the CNN pipeline, we used ANTS as opposed to FSL FLIRT for registration as it showed better AC-PC alignment.

## 4. Experiments and results

Both DCBO and SABO are applied to multiple predictive tasks with multiple inputs, including regression and The subnetworks include DFNNs and CNNs. As our methods are combined with BO, BO acts as the baseline comparison. For comparison between methods, regret is reported [9], defined at each timepoint $t$ as the difference between the best overall performance, and the best performance at time t. This is $min_t(L) - min(L)$ for regressive models when MSE is used as the loss for HPO, and $max(L) - max_t(L)$ for classification models when accuracy is used as the loss for HPO. Speedup is calculated at each time point as the minimum time the comparative methods (BO) took to reach regret G at time t, divided by the minimum time our method took to reach the same regret G. Therefore, speedup values over 1 indicate our method is faster. For clarity, the experiments in this work are split into groups, and summarized in table 1. All models are trained using the ADAM optimizer, and early stopping is used with a patience of 10 epochs and a max number of 500 epochs. Each trial was run multiple times to allow for a statistical comparison between methods. A performance summary (table 2) reports the mean speedup, max speedup, final speedup and final performance gain for each experiment. Mean speedup is defined as the mean over all time points, max speedup is the maximum seen over the experiment, and final speedup is computed at the end of the run. Gain in performance is the final gain in performance of our method vs BO at the end of the run. Representative regret plots are shown in fig. 5, and performance gain is shown in (fig. S2). Hyperparameter ranges for all experiments and are provided in the Supplemental Tables 1, 2 and 3.

### 4.1 DCBO experimental results

#### 4.1.1 Experiment: DCBO vs BO for image classification using CNN subnetworks (E$_{DCCC}$).

This set of experiments compares DCBO to BO with different numbers of inputs, and different amount of data. To achieve these conditions, each CIFAR image is split spatially into 3, 4, or 9 inputs (fig. 6a), and 20%, 50%, or 100% of the training data is used. CNN models are used for each subnetwork, except the dense merge network, which combines the flattened latent representation from each of the subnetworks to make a final prediction from all inputs. Categorical cross-entropy is used for model loss ($l$), and 1-validation accuracy is used for TPE loss (TPE always minimizes loss). Four trials are run for both DCBO and BO, with different initial random seeds. Mean, max and final speed up, along with final gain in performance (table 3), show that DCBO consistently outperforms BO with differing numbers of inputs and amounts of training data, with a speedup of up to 15.78x and final accuracy gain of up to 3.5%.

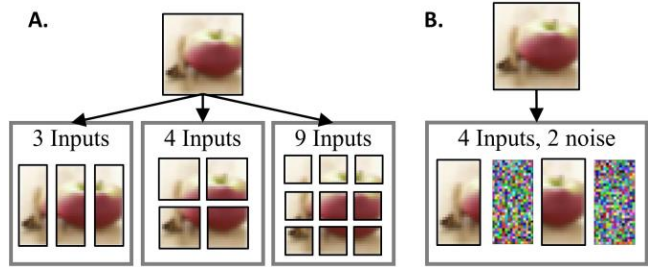

**Fig. 6.** CIFAR for DCBO experiments. Example image split into 3,4 and 9 inputs. **B.** CIFAR for IABO experiments. Example image split into 2 inputs, and 2 other noisy inputs.

**4.1.2 Experiment: DCBO vs BO for regression of future ADAS13 with DFNN subnetworks ($E_{DARD}$).**

This experiment also compares DCBO vs BO. However, inputs are tabular summary metrics from both clinical data and imaging data, and the task is to predict ADAS13 36 months in the future from time of measurment. Summary imaging metrics include regional volumes from T1 and mean regional SUVr from PET-FDG and PET-AV45. The network therefore has 4 DFNN input subnetworks, and 1 DFNN merge subnetwork. Ten-fold cross validation was carried out, grouped on subject and stratified on final ADAS13, age, and sex. Each fold acts as a pairwise comparison between DCBO and BO. The loss function ($l$) of the models is MSE, and validation MSE is used for the TPE loss function to optimize the configuration. DCBO showed a speedup of up to 23.62x and a final MSE improvement of 3.62.

**4.1.3 Experiment: DCBO vs BO for classification of MCI conversion with and DFNN subnetworks ($E_{DACD}$).**

Similar to $E_{DARD}$, this experiment compares DCBO vs BO. The inputs are the same as experiment $E_{DARD}$ using baseline tabular metrics form imaging and clinical data. The general model architecture is also similar, using DFNN as subnetworks. However, this experiment is classification, and the target is whether or not the subjects will convert to AD in 36 months (MCIc), or stay as stable MCI (MCIs). Ten-fold cross validation was also carried out, grouped on subject, and stratified on conversion, age, and sex. Each fold acts as a pairwise comparison between DCBO and BO. Categorical cross-entropy is used for model loss ($l$), and 1-validation accuracy is used for TPE loss. This experiment shows a speedup of up to 5.72x and a final accuracy gain of 1.1%.

**4.1.4 Experiment: DCBO vs BO for regression of future ADAS13 with CNN subnetworks ($E_{DACC}$).**

Our last DCBO vs BO experiment uses the ADNI dataset and predicts future ADAS13 score from 3D imaging data, rather than summary values. This experiment uses a general model architecture with three 3D CNNs as subnetworks, one each for the T1, FDG-PET, and AV45-PET, and a DFNN for the subnetwork for the clinical data. The merge network is a DFNN and concatenates the latent representations of the prior networks. Eight-fold cross validation was carried out, grouped on subject, and stratified on final ADAS13, age, and sex. Each fold acts as a pairwise comparison between DCBO and BO. The loss function ($l$) of the models, is MSE, and validation MSE is used for the TPE loss function to optimize the configuration. The maximum speedup achieved was 10.87x with a final MSE boost of 4.44.

**4.2 SABO experimental results**

**4.2.1 SABO vs BO for image classification using CNN subnetworks ($E_{ACCC}$)**

This experiment compares SABO and BO using the CIFAR dataset. To simulate multiple inputs of varying importance, CIFAR images are split into two, and two uniform random noise inputs of the same dimension are used to represent inputs with no signal (fig. 6b), and therefore less important corresponding subnetworks. Inputs 1 and 3 contain each half of CIFAR, and inputs 2 and 4 contain the noisy inputs. For this experiment, 5 initial complete models are trained, and the performance of each input's subnetworks are compared (sup fig. 1B). This is then used to detect which inputs are less important. SABO is then used to train new networks, while using the best prior network for the inputs that do not contain signal. Each method was run for 6 independent trials. This experiment showed a speedup of up to 4.08x and final accuracy increase of 1.8%. Similar to the DCBO CIFAR experiments, categorical cross-entropy is used for model loss ($l$), and 1-validation accuracy is used for TPE loss.

### 4.2.2 Experiment: SABO vs BO for regression of future ADAS13 with DFNN subnetworks ($E_{AARD}$)

Experiment $E_{AARD}$ compared SABO and BO on the ADNI dataset. Inputs to the subnetworks are tabular data from regional T1 intensity from cortical white matter (as opposed to volume used in DCBO), and regional mean SUVr from FDG-PET. The T1 data is intentionally limited to provide an input with less signal and a less important subnetwork. Ten-fold cross validation was carried out, grouped on subject, and stratified on final ADAS13, age, and sex. The loss function ($l$) of the models is MSE, and validation MSE is used for the TPE loss function to optimize the configuration. A maximum speedup of 6.48x was achieved and a final MSE boost of 3.22.

### 4.2.3 Experiment: SABO vs BO for classification of MCI conversion with and DFNN subnetworks ($E_{AACD}$)

Our final experiment is similar to $E_{AARD}$. The general model architecture, inputs, train/val data split, and data remains consistent. However, this experiment uses classification models, and the target is MCIc vs MCIs. This experiment showed a max speed up of 3.66x and a final boost in accuracy of 1.1% basis points.

## 5. Discussion.

In this manuscript we describe two new hyperparameter approaches, divide-and-conquer (DC) and subnetwork adaptive (SA) HPO for the optimization of networks with multi-subnetwork architecture. They are combined with the commonly-used Bayesian optimization to create divide-and-conquer Bayesian optimization (DCBO) and subnetwork adaptive Bayesian optimization (SABO). Subsequently, they are compared to conventional BO across many different experiments, consisting of multiple datasets and different general architectures. Our approaches are consistently faster than the BO method by up to 23.6x. Additionally, in multiple experiments, the final speedup is at the end or close to the end of the run, and performance differences are increasing. This suggests that if the experiments ran for a longer period of time, even higher speed ups may be observed.

While this manuscript focuses on comparisons with BO, nearly all other general methods, as described in the introduction, do not leverage the separability of subnetworks though transfer learning. Our approaches are readily adaptable to other methods and will likely provide additional efficiencies when being applied to multi-subnetwork neural networks. Additionally, this manuscript focused on a commonly used, but specific type of multi-subnetwork model: multi-modal architectures. However, these methods are applicable to other multi-subnetwork architectures, as described in the introduction, as long as performance metrics can be calculated for each subnetwork. However, it is likely, that the use of transfer learning alone (fig. 7) is enough to provide increase efficiencies for HPO, and thus performance metrics may not be required for subnetworks.

To help determine the cause of the speed up for DCBO, we looked at the training time and number of epochs for experiment $E_{DARC}$ (fig. 7). Overall, this experiment showed a final speed up of 10.87x. However, when comparing the total training time per model for DCBO vs BO, the average training time for BO models is 516 seconds, and for DCBO is 201 seconds (many of which are transfer models). We also see a very small increase in the number of epochs from 27 to 29 from BO to DCBO, which is likely due to the fact that the models are learning more and thus require additional training iterations. Combined, these suggest that transfer learning is providing ~2.5x speed up, and the additional ~4x (2.5*4=10) is most likely from intelligent combination and early application of BO to the subnetworks. While ideally HPO methods that embed DC would be capable of leveraging both transfer learning, and intelligent combination and early optimization of the subnetworks, this demonstrates that both aspects offer speedups, and other existing HPO methods need only to take advantage of one to offer a significant speed up.

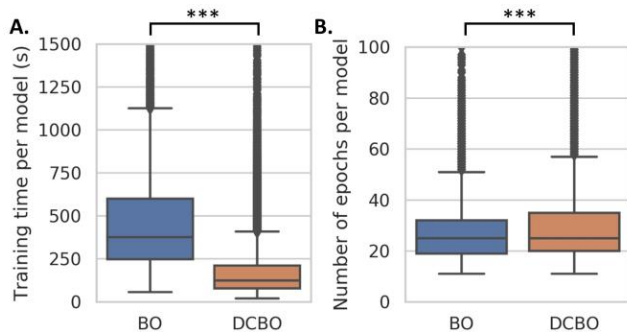

**Fig 7**. Training time and number of training epochs for $E_{DARC}$: A. Box plot showing the training time per model split by method. B. Box plot showing the number of epochs per model split by method. Both plots cropped to highlight the non-outlier differences in training time and number of epochs. Both p-values are highly significant $p < 1e\text{-}38$.

In this work we calculate speed up as the minimum time our method took to reach a regret G / the minimum time the comparative method (BO) to reach the same R. This is a conservative approach, as it does not take into account the time BO may have been stuck at a specific regret that our approaches quickly outperform. If instead we calculated regret as the minimum time our method took to have a lower a regret than G / the minimum time the comparative method (BO) to reach G, some speed ups would have been significantly larger. E.g. $E_{DACD}$ has a final speed up of 1.8x, however if we calculated as above, we would see a final speed up of 86400/19000 = 4.55x. However, to help ensure the validity of our results, and to facilitate comparison to other HPO manuscripts, we selected the conservative approach. Similarly, we focus on the optimization of validation performance, as this is the most common approach for hyperparameter optimization.

Lastly, while we provide and test 2 approaches in this work, they are not mutually exclusive and could likely be combined and provide for additional optimization efficiency. To do so, the probability of applying transfer learning in DC could be modified per network based on the approach in SA. However, as both our methods are gaining speed ups from the general approaches in part due to transfer learning, and the inclusion of SA in DC would simply modify the likelihood of using transfer learning, we do not expect that the speed up over general approaches would be additive. Rather, the combination of DC and SA would provide a moderate speedup over DC, however the magnitude would have to be determined empirically and is probably highly dependent on the importance of the subnetworks for performance (e.g. if all subnetworks are of equal importance, the combination would likely not improve efficacy over just applying a DC approach). Also, while DC and SA both demonstrate the ability to leverage subnetwork architecture for hyperparameter optimization, our largest speed ups and highest performance gains are achieved with DC. Additionally, as DC also makes no assumptions about relative performance, we suggest DC to be the primary HPO method to tested out of the two.

## 5. Limitations

aOur work focuses on a subset of subnetwork architectures, which includes CNN and DFNN. While we do expect that our methods would work well on other types of networks, including RNN, LSTM, autoencoders, etc., further testing would be required to validate the application of DCBO and SABO to these networks. We also only integrate and compare our methods to BO. Again, we expect that they will provide speed ups to other HPO methods, but empirical testing is required to determine if the increase in efficiency is similar to the results presented here. Lastly, we only tested our approaches on Nvidia Tesla p100 and v100 GPUs, while using TensorFlow. While this is a very common setup, other deep learning frameworks may not handle transfer learning and/or freezing weights as well as TensorFlow, thus decreasing the efficiency gained from transfer learning.

## 6. Conclusion

This work describes two novel approaches for the efficient hyperparameter optimization of deep learning models with multi-subnetwork architectures and integrates them both with a commonly used HPO method, BO. In our experiments, these methods consistently outperform BO, which does not leverage multi-subnetwork architecture. We consistently show reduced computational resources to achieve the same performance, up to 23x, across both practical and synthetic predictive tasks using multiple different subnetwork model types including DFNN and CNNs. Therefore, we suggest these methods should be employed when optimizing complex deep learning models that consist of multiple separable subnetworks.

## 7. Code availability:

Code is made available for academic use: link to git

## 8. Acknowledgements

The ADNI data used in this article was funded by the Alzheimer's Disease Neuroimaging Initiative (ADNI) (National Institutes of Health Grant U01 AG024904) and DOD ADNI (Department of Defense award number W81XWH-12-2-0012). ADNI is funded by the National Institute on Aging, the National Institute of Biomedical Imaging and Bioengineering, and through generous contributions from the following: AbbVie, Alzheimer's Association; Alzheimer's Drug Discovery Foundation; Araclon Biotech; BioClinica, Inc.; Biogen; Bristol-Myers Squibb Company; CereSpir, Inc.; Cogstate; Eisai Inc.; Elan Pharmaceuticals, Inc.; Eli Lilly and Company; EuroImmun; F. Hoffmann-La Roche Ltd and its affiliated company Genentech, Inc.; Fujirebio; GE Healthcare; IXICO Ltd.;Janssen Alzheimer Immunotherapy Research & Development, LLC.; Johnson & Johnson Pharmaceutical Research & Development LLC.; Lumosity; Lundbeck; Merck & Co., Inc.;Meso Scale Diagnostics, LLC.; NeuroRx Research; Neurotrack Technologies; Novartis Pharmaceuticals Corporation; Pfizer Inc.; Piramal Imaging; Servier; Takeda Pharmaceutical Company; and Transition Therapeutics. The Canadian Institutes of Health Research is providing funds to support ADNI clinical sites in Canada. Private sector contributions are facilitated by the Foundation for the National Institutes of Health (www.fnih.org). The grantee organization is the Northern California Institute for Research and Education, and the study is coordinated by the Alzheimer's Therapeutic Research Institute at the University of Southern California. ADNI data are disseminated by the Laboratory for Neuro Imaging at the University of Southern California.

## 9. References

[1] G. E. Hinton, "A Practical Guide to Training Restricted Boltzmann Machines," in *LNCS sublibrary. SL 1, Theoretical computer science and general issues*, vol. 7700, *Neural networks: Tricks of the trade*, G. Montavon, G. Orr, and K.-R. Müller, Eds., 2nd ed., Heidelberg: Springer, 2012, pp. 599–619.

[2] J. Bergstra and Y. Bengio, "Random Search for Hyper-Parameter Optimization," *Journal of Machine Learning Research*, vol. 13, pp. 281–305, 2012.

[3] T. Chen, I. Goodfellow, and Shlens J., "Net2Net: Accelerating learning via knowledge transfer," *International Conference on Learning Representations*, 2016.

[4] M. Jaderberg *et al.,* "Population Based Training of Neural Networks," Nov. 2017. [Online]. Available: http://arxiv.org/pdf/1711.09846v2

[5] J. Bergstra, R. Bardenet, Y. Bengio, and B. Kégl, "Algorithms for Hyper-Parameter Optimization," *Conference proceedings : Neural Information Processing Systems 2011*, pp. 2546–2554, 2011.

[6] L. Li, K. Jamieson, G. DeSalvo, A. Rostamizadeh, and A. Talwalkar, "Hyperband: A Novel Bandit-Based Approach to Hyperparameter Optimization," *Journal of Machine Learning Research*, 2018. [Online]. Available: http://arxiv.org/pdf/1603.06560v4

[7] B. Zoph and Q. V. Le, "Neural architecture search with reinforcement learning," *ICLR*, 2017.

[8] H. Pham, M. Y. Guan, B. Zoph, Le Q. V., and Dean J., "Efficient neural architecture search via parameter sharing," *ICML*, 2018. [Online]. Available: http://arxiv.org/pdf/1802.03268v2

[9] S. Falkner, A. Klein, and F. Hutter, "BOHB: Robust and Efficient Hyperparameter Optimization at Scale," *International Conference on Machine Learning*, vol. 4, pp. 2323–2341, 2018. [Online]. Available: http://arxiv.org/pdf/1807.01774v1

[10] K. Noda, Y. Yamaguchi, K. Nakadai, H. G. Okuno, and T. Ogata, "Audio-visual speech recognition using deep learning," *Appl Intell*, vol. 42, no. 4, pp. 722–737, 2015, doi: 10.1007/s10489-014-0629-7.

[11] J. Shi, X. Zheng, Y. Li, Q. Zhang, and S. Ying, "Multimodal Neuroimaging Feature Learning With Multimodal Stacked Deep Polynomial Networks for Diagnosis of Alzheimer's Disease," *IEEE journal of biomedical and health informatics*, vol. 22, no. 1, pp. 173–183, 2018, doi: 10.1109/JBHI.2017.2655720.

[12] S. Aslan, U. Güdükbay, and H. Dibeklioğlu, "Multimodal assessment of apparent personality using feature attention and error consistency constraint," *Image and Vision Computing*, vol. 110, p. 104163, 2021, doi: 10.1016/j.imavis.2021.104163.

[13] X. Wu, W.-L. Zheng, Z. Li, and B.-L. Lu, "Investigating EEG-based functional connectivity patterns for multimodal emotion recognition," *Journal of neural engineering*, vol. 19, no. 1, 2022, doi: 10.1088/1741-2552/ac49a7.

[14] H. Sharifi-Noghabi, O. Zolotareva, C. C. Collins, and M. Ester, "MOLI: multi-omics late integration with deep neural networks for drug response prediction," *Bioinformatics (Oxford, England)*, vol. 35, no. 14, i501-i509, 2019, doi: 10.1093/bioinformatics/btz318.

[15] D. Hong, L. Gao, X. Wu, J. Yao, N. Yokoya, and B. Zhang, "A Unified Multimodal Deep Learning Framework For Remote Sensing Imagery Classification," in *2021 11th Workshop on Hyperspectral Imaging and Signal Processing: Evolution in Remote Sensing (WHISPERS)*, Amsterdam, Netherlands, Mar. 2021 - Mar. 2021, pp. 1–5.

[16] X. Zhao, Y. Wu, G. Song, Z. Li, Y. Zhang, and Y. Fan, "A deep learning model integrating FCNNs and CRFs for brain tumor segmentation," *Medical image analysis*, vol. 43, pp. 98–111, 2018, doi: 10.1016/j.media.2017.10.002.

[17] J. Wu *et al.,* "Deep morphology aided diagnosis network for segmentation of carotid artery vessel wall and diagnosis of carotid atherosclerosis on black-blood vessel wall MRI," *Medical physics*, vol. 46, no. 12, pp. 5544–5561, 2019, doi: 10.1002/mp.13739.

[18] K. Gu, Z. Xia, J. Qiao, and W. Lin, "Deep Dual-Channel Neural Network for Image-Based Smoke Detection," *IEEE Trans. Multimedia*, vol. 22, no. 2, pp. 311–323, 2020, doi: 10.1109/TMM.2019.2929009.

[19] Y.-C. Chiu *et al.,* "Predicting drug response of tumors from integrated genomic profiles by deep neural networks," *BMC Med Genomics*, vol. 12, Suppl 1, p. 18, 2019, doi: 10.1186/s12920-018-0460-9.

[20] J.-M. Perez-Rua, V. Vielzeuf, S. Pateux, M. Baccouche, and F. Jurie, "MFAS: Multimodal Fusion Architecture Search," in *2019 IEEE/CVF Conference on Computer Vision and Pattern Recognition (CVPR)*, Long Beach, CA, USA, 2019, pp. 6959–6968.

[21] Z. Yu, Y. Cui, J. Yu, M. Wang, D. Tao, and Q. Tian, "Deep Multimodal Neural Architecture Search," in *Proceedings of the 28th ACM International Conference on Multimedia*, Seattle WA USA, 2020, pp. 3743–3752.

[22] J. Snoek, H. Larochelle, and R. P. Adams, "Practical Bayesian Optimization of Machine Learning Algorithms," *Advances in Neural Information Processing Systems*, vol. 25, 2012.

[23] G. van Rossum and F. L. Drake Jr, *Python reference manual*: Centrum voor Wiskunde en Informatica Amsterdam, 1995.

[24] F. Chollet and et. al, *Keras*.

[25] Martín Abadi *et al., TensorFlow: Large-Scale Machine Learning on Heterogeneous Systems.* [Online]. Available: https://www.tensorflow.org/

[26] C. R. Harris *et al.,* "Array programming with NumPy," *Nature*, vol. 585, no. 7825, pp. 357–362, 2020, doi: 10.1038/s41586-020-2649-2.

[27] W. McKinney, "Data Structures for Statistical Computing in Python," in *Proceedings of the 9th Python in Science Conference*, Austin, Texas, 2010, pp. 56–61.

[28] Fabian Pedregosa *et al.,* "Scikit-learn: Machine Learning in Python," *Journal of Machine Learning Research*, vol. 12, no. 85, pp. 2825–2830, 2011. [Online]. Available: http://jmlr.org/papers/v12/pedregosa11a.html

[29] O. Tange, *Gnu Parallel 2018*: Zenodo, 2018.

[30] B. Fischl, "FreeSurfer," *NeuroImage*, vol. 62, no. 2, pp. 774–781, 2012, doi: 10.1016/j.neuroimage.2012.01.021.

[31] M. W. Woolrich *et al.,* "Bayesian analysis of neuroimaging data in FSL," *NeuroImage*, vol. 45, 1 Suppl, S173-86, 2009, doi: 10.1016/j.neuroimage.2008.10.055.

[32] P. Moritz *et al.,* "Ray: A Distributed Framework for Emerging AI Applications," Dec. 2017. [Online]. Available: http://arxiv.org/pdf/1712.05889v2

[33] A. Krizhevsky and G. Hinton, *Learning multiple layers of features from tiny images*, 2009. [Online]. Available: http://citeseerx.ist.psu.edu/viewdoc/download?doi=10.1.1.222.9220&rep=rep1&type=pdf

[34] R. C. Petersen *et al.,* "Alzheimer's Disease Neuroimaging Initiative (ADNI): clinical characterization," *Neurology*, vol. 74, no. 3, pp. 201–209, 2010, doi: 10.1212/WNL.0b013e3181cb3e25.

[35] O. Lucena, R. Souza, L. Rittner, R. Frayne, and R. Lotufo, "Convolutional neural networks for skull-stripping in brain MR imaging using silver standard masks," *Artificial intelligence in medicine*, vol. 98, pp. 48–58, 2019, doi: 10.1016/j.artmed.2019.06.008.

[36] G. Grabner, A. L. Janke, M. M. Budge, D. Smith, J. Pruessner, and D. L. Collins, "Symmetric atlasing and model based segmentation: an application to the hippocampus in older adults," *Medical image computing and computer-assisted intervention : MICCAI ... International Conference on Medical Image Computing and Computer-Assisted Intervention*, vol. 9, Pt 2, pp. 58–66, 2006, doi: 10.1007/11866763_8.